\pdfoutput=1

\documentclass[11pt]{article}

\usepackage[]{emnlp2021}

\usepackage{times}
\usepackage{latexsym}

\usepackage[T1]{fontenc}

\usepackage[utf8]{inputenc}

\usepackage{microtype}

\usepackage{graphicx}
\usepackage{eurosym}
\usepackage{tablefootnote}
\usepackage{kantlipsum}
\usepackage{tabularx}
\graphicspath{ {images/} }
\usepackage{hyperref}
\usepackage{amsmath}
\usepackage{hhline}
\usepackage[utf8]{inputenc}
\usepackage{multirow}

\usepackage[T1]{fontenc}
\usepackage{kantlipsum}
\usepackage{listings}
\usepackage{stfloats}
\usepackage{csquotes}
\usepackage{array}
\usepackage{booktabs}
\usepackage[super]{nth}
\usepackage[inline,shortlabels]{enumitem}
\usepackage{enumitem}
\usepackage{makecell}
\usepackage{siunitx}
\usepackage{xcolor, colortbl}
\usepackage{bm}
\usepackage{url}
\usepackage{wrapfig}
\usepackage{amsmath}
\usepackage{amssymb}
\usepackage{mathtools}
\usepackage{amsfonts}
\usepackage{enumitem}
\usepackage{graphicx}
\usepackage{subcaption}
\usepackage{nicefrac}       
\usepackage{tablefootnote}
\usepackage{bbm}
\usepackage{microtype} 
\usepackage{dsfont}
\usepackage{lipsum,changepage}

\usepackage{algorithm} 
\usepackage[algo2e,ruled,linesnumbered]{algorithm2e} 
\usepackage[]{algpseudocode}

\newcommand{\entropy}{\textsc{Entropy}}

\newcommand{\rand}{\textsc{Random}}
\newcommand{\alps}{\textsc{Alps}}
\newcommand{\badge}{\textsc{Badge}}
\newcommand{\bkm}{\textsc{BertKM}}
\newcommand{\call}{\textsc{Cal}}

\newcommand{\bert}{\textsc{Bert}}

\newcommand{\Dpool}{\mathcal{D}_{\textbf{pool}}}
\newcommand{\Dlab}{\mathcal{D}_{\textbf{lab}}}

\newcommand{\cls}{\texttt{[CLS]}}

\newcommand{\imdb}{\textsc{imdb}}
\newcommand{\sst}{\textsc{sst-$2$}}
\newcommand{\qnli}{\textsc{qnli}}
\newcommand{\qqp}{\textsc{qqp}}
\newcommand{\dbpedia}{\textsc{dbpedia}}
\newcommand{\ag}{\textsc{agnews}}
\newcommand{\pub}{\textsc{pubmed}}
\newcommand{\twit}{\textsc{twitterppdb}}

\title{Active Learning by Acquiring Contrastive Examples}

\author{Katerina Margatina$^\dagger$ \enspace 
Giorgos Vernikos$^{\ddagger\ast}$ \enspace
 Lo\"{i}c Barrault$^\dagger$ \enspace 
 Nikolaos Aletras$^\dagger$ \\
    $^\dagger$University of Sheffield \enspace
    $^\ddagger$EPFL \enspace
        $^\ast$HEIG-VD\\
  \texttt{\{k.margatina, l.barrault, n.aletras\}@sheffield.ac.uk} \\
  \texttt{georgios.vernikos@epfl.ch}
  }

\begin{document}
\maketitle
\begin{abstract}

Common acquisition functions for active learning use either uncertainty or diversity sampling, aiming to select difficult and diverse data points from the pool of unlabeled data, respectively. In this work, leveraging the best of both worlds, we propose an acquisition function that opts for selecting \textit{contrastive examples}, i.e. data points that are similar in the model feature space and yet the model outputs maximally different predictive likelihoods. We compare our approach, \call{} (Contrastive Active Learning),  with a diverse set of acquisition functions in four natural language understanding tasks and seven datasets.
Our experiments show that \call{} performs consistently better or equal than the best performing baseline across all tasks, on both in-domain and out-of-domain data. 
We also conduct an extensive ablation study of our method and we further analyze all actively acquired datasets showing that \call{} achieves a better trade-off between uncertainty and diversity compared to other strategies.

\end{abstract}

\section{Introduction}\label{sec:intro}
Active learning (AL) is a machine learning paradigm for efficiently acquiring data for annotation from a (typically large) pool of unlabeled data~\cite{Lewis1994-mj, Cohn:1996:ALS:1622737.1622744,settles2009active}.
Its goal is to concentrate the human labeling effort on the most informative data points that will benefit model performance the most and thus reducing data annotation cost.

The most widely used approaches to acquiring data for AL are based on uncertainty and diversity, often described as the ``two faces of AL''~\cite{DASGUPTA20111767}. 
While uncertainty-based methods leverage the model predictive confidence to select difficult examples for annotation~\cite{Lewis:1994:SAT:188490.188495,Cohn:1996:ALS:1622737.1622744}, diversity sampling exploits heterogeneity in the feature space by typically performing clustering~\cite{Brinker03incorporatingdiversity, pmlr-v16-bodo11a}.
Still, both approaches have core limitations that may lead to acquiring redundant data points. Algorithms based on uncertainty may end up choosing uncertain yet uninformative repetitive data, while diversity-based methods may tend to select diverse yet easy examples for the model~\cite{10.5555/645530.655646}.
The two approaches are orthogonal to each other, since uncertainty sampling is usually based on the model's output, while diversity exploits information from the input (i.e. feature) space. Hybrid data acquisition functions that combine uncertainty and diversity sampling have also been proposed \cite{shen-etal-2004-multi, zhu-etal-2008-active,
Ducoffe2018-sq,
Ash2020Deep,yuan-etal-2020-cold,
ru-etal-2020-active}.

\begin{figure}[!t]
        \centering
        \includegraphics[width=0.7\columnwidth]{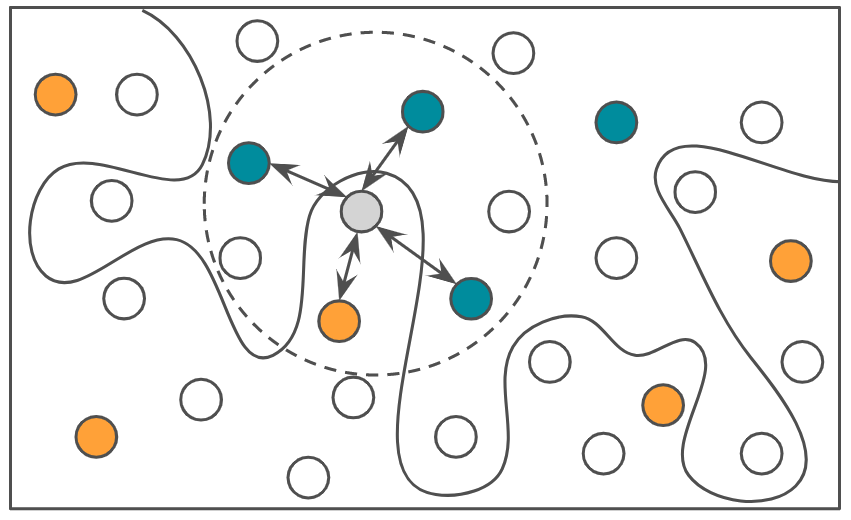}

    \caption{Illustrative example of our proposed method \call{}. The solid line (model decision boundary) separates data points from two different classes (blue and orange), the coloured data points represent the labeled data and the rest are the unlabeled data of the pool. 
    }
    \label{fig:toy}
\end{figure}
In this work, we aim to leverage characteristics from hybrid data acquisition. We hypothesize that data points that are close in the model feature space (i.e. share similar or related vocabulary, or similar model encodings) but the model produces different predictive likelihoods, should be good candidates for data acquisition. We define such examples as \textit{contrastive} (see example in Figure~\ref{fig:toy}). 
For that purpose, we propose a new acquisition function that searches for contrastive examples in the pool of unlabeled data.
Specifically, our method, Contrastive Active Learning (\call{}) \textit{selects unlabeled data points from the pool, whose predictive likelihoods diverge the most from their neighbors in the training set}. This way, \call{} shares similarities with diversity sampling, but instead of performing clustering it uses the feature space to create neighborhoods. \call{} also leverages uncertainty, by using predictive likelihoods to rank the unlabeled data.

We evaluate our approach in seven datasets from four tasks including sentiment analysis, topic classification, natural language inference and paraphrase detection. We compare \call{} against a full suite of baseline acquisition functions that are based on uncertainty, diversity or both. We also examine robustness by evaluating on out-of-domain data, apart from in-domain held-out sets. Our contributions are the following:
\begin{enumerate}
    \item We propose \call{}, a new acquisition function for active learning that acquires contrastive examples from the pool of unlabeled data (\S\ref{sec:method});
    \item We show that \call{} performs consistently better or equal compared to all baselines in all tasks when evaluated on in-domain and out-of-domain settings (\S\ref{sec:results});
    \item We conduct a thorough analysis of our method showing that \call{} achieves a better trade-off between diversity and uncertainty compared to the baselines (\S\ref{sec:analysis}).
\end{enumerate}
We release our code online~\footnote{\url{https://github.com/mourga/contrastive-active-learning}}.

\section{Contrastive Active Learning}\label{sec:method}
In this section we present in detail our proposed method, \call{}: Contrastive Active Learning. First, we provide a definition for contrastive examples and how they are related to finding data points that are close to the decision boundary of the model (\S\ref{sec:contr_ex}). 
We next describe an active learning loop using our proposed acquisition function (\S\ref{sec:al_loop}).

\subsection{Contrastive Examples}\label{sec:contr_ex}
In the context of active learning, we aim to formulate an acquisition function that selects contrastive examples from a pool of unlabeled data for annotation.
We draw inspiration from the contrastive learning framework, that leverages the similarity between data points to push those from the same class closer together and examples from different classes further apart during training~\cite{10.5555/2999792.2999959,NIPS2016_6b180037,oord2019representation,pmlr-v119-chen20j,gunel2021supervised}.

In this work, we define as contrastive examples 
two data points if their model encodings are similar, but their model predictions are very different (maximally disagreeing predictive likelihoods).

Formally, data points
$x_i$ and $x_j$ should first satisfy a similarity criterion:
\begin{equation}\label{eq:prop1}
    d\big(\Phi(x_i),\Phi(x_j)\big)<\epsilon
\end{equation}
where $\Phi(.) \in \mathbb{R}^{d^{'}}$ is an encoder that maps $x_i, x_j$ in a shared feature space, $d(.)$ is a distance metric and $\epsilon$ is a small distance value. 

A second criterion, based on model uncertainty, is to evaluate
that the predictive probability distributions of the model $p(y|x_i)$ and $p(y|x_j)$ for the inputs $x_i$ and $x_j$ should maximally diverge:
\begin{equation}\label{eq:prop2}
    \mathrm{KL}\big(p(y|x_i)|| p(y|x_j)\big)
    \rightarrow \infty
\end{equation}
where KL is the Kullback-Leibler divergence between two probability distributions~\footnote{KL divergence is not a symmetric metric, $\mathrm{KL}(P||Q)=\sum\limits_{x} P(x)\mathrm{log}\big(\frac{P(x)}{Q(x)}\big)$. We use as input $Q$ the output probability distribution of an unlabeled example from the pool and as target $P$ the output probability distribution of an example from the train set (See \S\ref{sec:al_loop} and algorithm~\ref{algo:contrastive}).}.
%

For example, in a binary classification problem, given a reference example $x_1$ with output probability distribution ($0.8,0.2$)~\footnote{A predictive distribution ($0.8,0.2$) here denotes that the model is $80\%$ confident that $x_1$ belongs to the first class and $20\%$ to the second.} and similar candidate examples $x_2$ with ($0.7, 0.3$) and $x_3$ with ($0.6, 0.4$), we would consider as contrastive examples the pair $(x_1, x_3)$.
However, if another example $x_4$ (similar to $x_1$ in the model feature space) had a probability distribution ($0.4, 0.6$), then the most contrastive pair would be ($x_1$, $x_4$).

Figure~\ref{fig:toy} provides an illustration of contrastive examples for a binary classification case. All data points inside the circle (dotted line) are similar in the model feature space, satisfying Eq.~\ref{eq:prop1}. Intuitively, if the divergence of the output probabilities of the model for the gray and blue shaded data points is high, then Eq.~\ref{eq:prop2} should also hold and we should consider them as contrastive. 
   
From a different perspective, 
data points with similar model encodings (Eq.~\ref{eq:prop1}) and dissimilar model outputs (Eq.~\ref{eq:prop2}), should be close to the model's decision boundary (Figure~\ref{fig:toy}).
Hence, we hypothesize that our proposed approach to select contrastive examples is related to acquiring difficult examples near the decision boundary of the model.
Under this formulation, \call{} does not guarantee that the contrastive examples lie near the model's decision boundary, because our definition is not strict. In order to ensure that a pair of contrastive examples lie on the boundary, the second criterion should require that the model classifies the two examples in different classes (i.e. different predictions). However, calculating the distance between an example and the model decision boundary is intractable and approximations that use adversarial examples are computationally expensive~\cite{Ducoffe2018-sq}.

\begin{algorithm*}[t!]
	\caption{Single iteration of \textsc{cal} \label{algo:contrastive}}

    \DontPrintSemicolon
    \SetAlgoLined

	\KwIn{labeled data $\Dlab$, unlabeled data $\Dpool$, acquisition size $b$, model $\mathcal{M}$, number of neighbours $k$, model representation (encoding) function $\Phi(.)$}
	\For {$x_p$ in $\Dpool$}{
	$ \big\{(x_l^{(i)},y_l^{(i)})\big\}, i=1,...,k \leftarrow \mathrm{KNN}\big(\Phi(x_p), \Phi(\Dlab), k\big)$  \Comment{find neighbours in $\Dlab$}\;
	
	$p(y|x_l^{(i)})    \leftarrow \mathcal{M}(x_l^{(i)}), i=1,...,k$ \Comment{compute probabilities}\;
	
	$p(y|x_p)  \leftarrow \mathcal{M}(x_p)$ \;
	
	$\mathrm{KL}\big(p(y|x_l^{(i)})||p(y|x_p)\big), i=1,...,k$ \Comment{compute divergence}\;
	
	$s_{x_p} = \frac{1}{k} \sum\limits_{i=1}^k \mathrm{KL}\big(p(y|x_l^{(i)})||p(y|x_p)\big)$ \Comment{score}\;
	}
	
	$ Q =\underset{x_p \in \Dpool}{\mathrm{argmax }} s_{x_p}, |Q|=b $\Comment{select batch}\;

	\KwOut{ $Q$ }
\end{algorithm*}


\subsection{Active Learning Loop}\label{sec:al_loop}
Assuming a multi-class classification problem with $C$ classes, labeled data for training $\Dlab$ and a pool of unlabeled data $\Dpool$,
we perform AL for $T$ iterations. At each iteration, we train a model on $\Dlab$ and then use our proposed acquisition function, \call{} (Algorithm~\ref{algo:contrastive}), to acquire a batch $Q$ consisting of $b$ examples from $\Dpool$.
The acquired examples are then labeled\footnote{We simulate AL, so we already have the labels of the examples of $\Dpool$ (but still treat it as an unlabeled dataset).}, they are removed from the pool $\Dpool$ and added to the labeled dataset $\Dlab$, which will serve as the training set for training a model in the next AL iteration.
In our experiments, we use a pretrained \bert{} model $\mathcal{M}$~\cite{Devlin2019-ou}, which we fine-tune at each AL iteration using the current $\Dlab$.
We begin the AL loop by training a model $\mathcal{M}$ using an initial labeled dataset $\Dlab{}$~\footnote{We acquire the first examples that form the initial training set $\Dlab{}$ by applying random stratified sampling (i.e. keeping the initial label distribution).}.

\paragraph{Find Nearest Neighbors for Unlabeled Candidates} The first step of our contrastive acquisition function (cf. line 2) is to find examples that are similar in the model feature space (Eq.~\ref{eq:prop1}). Specifically, we use the \cls{} token embedding of \bert{} as our encoder $\Phi(.)$ to represent all data points in $\Dlab$ and $\Dpool$. We use a K-Nearest-Neighbors (KNN) implementation using the labeled data $\Dlab$, in order to query similar examples  $ x_l \in \Dlab$ for each candidate $x_p \in \Dpool$. Our distance metric $d(.)$ is Euclidean distance. To find the most similar data points in $\Dlab$ for each $x_p$, we select the top $k$ instead of selecting a predefined threshold $\epsilon$ (Eq.~\ref{eq:prop1})~\footnote{We leave further modifications of our scoring function as future work. One approach would be to add the average distance from the neighbors (cf. line 6) in order to alleviate the possible problem of selecting outliers.}. This way, we create a neighborhood $N_{x_p} = \big\{x_p, x_l^{(1)}, \dots, x_l^{(k)}\big\}$ that consists of the unlabeled data point $x_p$ and its $k$ closest examples $x_l$ in $\Dlab$ (Figure~\ref{fig:toy}).

\paragraph{Compute Contrastive Score between Unlabeled Candidates and Neighbors} In the second step, we compute the divergence in the model predictive probabilities for the members of the neighborhood (Eq. \ref{eq:prop2}). Using the current trained model $\mathcal{M}$ to obtain the output probabilities for all data points in $N_{x_p}$ (cf. lines 3-4), we then compute the Kullback–Leibler divergence (KL) between the output probabilities of $x_p$ and all $x_l \in N_{x_p}$ (cf. line 5).
To obtain a score $s_{x_p}$ for a candidate $x_p$, we take the average of all divergence scores (cf. line 6). 

\paragraph{Rank Unlabeled Candidates and Select Batch} 
We apply these steps to all candidate examples $x_p 
\in \Dpool$ and obtain a score $s_{x_p}$ for each.
With our scoring function we define as contrastive examples the unlabeled data $x_p$ that have the highest score $s_{x_p}$. 
A high $s_{x_p}$ score indicates that the unlabeled data point $x_p$ has a high divergence in model predicted probabilities compared to its neighbors in the training set (Eq.~\ref{eq:prop1},~\ref{eq:prop2}), suggesting that it may lie near the model's decision boundary.
To this end, our acquisition function selects the top $b$ examples from the pool that have the highest score $s_{x_p}$ (cf. line 8), that form the acquired batch $Q$.

\section{Experimental Setup}
\subsection{Tasks \& Datasets}
We conduct experiments on sentiment analysis, topic classification, natural language inference and paraphrase detection tasks. We provide details for the datasets in Table~\ref{table:datasets}. We follow \citet{yuan-etal-2020-cold} and use \imdb{}~\cite{maas-etal-2011-learning}, \sst{}~\cite{socher-etal-2013-recursive}, \pub{}~\cite{dernoncourt-lee-2017-pubmed} and \ag{} from ~\citet{NIPS2015_250cf8b5} where we also acquired \dbpedia{}. We experiment with tasks requiring pairs of input sequences, using \qqp{} and \qnli{} from \textsc{glue}~\cite{wang2018glue}.
To evaluate robustness on out-of-distribution (OOD) data, we follow \citet{hendrycks-etal-2020-pretrained} and use \sst{} as OOD dataset for \imdb{} and vice versa. We finally use \twit{}~\cite{lan-etal-2017-continuously} as OOD data for \qqp{} as in \citet{Desai2020-ys}.

\setlength{\tabcolsep}{6pt} 
\renewcommand{\arraystretch}{1.1} 

\begin{table*}[!t]
\resizebox{\textwidth}{!}{%
\centering
\begin{tabular}{lccccccc}
\Xhline{2\arrayrulewidth}

\textsc{dataset} & \textsc{task} & \textsc{domain} & \textsc{ood dataset} & \textsc{train} & \textsc{val}  & \textsc{test} & \textsc{classes}\\\hline 
\textsc{imdb}  & Sentiment Analysis & Movie Reviews & \textsc{sst-2} & $22.5$K & $2.5$K & $25$K & $2$\\
\textsc{sst-2} & Sentiment Analysis & Movie Reviews & \textsc{imdb} &$60.$6K  & $6.7$K & $871$& $2$\\
\textsc{agnews} & Topic Classification & News & - & $114$K  & $6$K & $7.6$K & $4$\\
\textsc{dbpedia} & Topic Classification & News & - & $20$K  & $2$K & $70$K & $14$\\
\textsc{pubmed} & Topic Classification & Medical & - & $180$K  & $30.2$K & $30.1$K & $5$\\
\textsc{qnli} & Natural Language Inference & Wikipedia & - & $99.5$K  &  $5.2$K & $5.5$K & $2$\\
\textsc{qqp} & Paraphrase Detection & Social QA Questions & \textsc{twitterppdb} & $327$K  & $36.4$K  & $80.8$K & $2$\\

\Xhline{2\arrayrulewidth}
\end{tabular}
}

\caption{Dataset statistics.}

\label{table:datasets}
\end{table*}
\subsection{Baselines} We compare \textsc{Cal} against five baseline acquisition functions. The first method,  \entropy{} is the most commonly used uncertainty-based baseline that acquires data points for which the model has the highest predictive entropy.
As a diversity-based baseline, following ~\citet{yuan-etal-2020-cold}, we use \bkm{} that applies k-means clustering using the $l_2$ normalized \bert{} output embeddings of the fine-tuned model to select $b$ data points.
We compare against
\badge{}~\cite{Ash2020Deep}, an acquisition function that aims to combine diversity and
uncertainty sampling, by computing \textit{gradient embeddings} $g_x$ for every candidate data point $x$ in $\Dpool$ and then using clustering to select a batch. Each $g_x$ is computed as the gradient of the cross-entropy loss with respect to the parameters of the model’s last layer, aiming to be the component that incorporates uncertainty in the acquisition function~\footnote{We note that \bkm{} and \badge{} are computationally heavy approaches that require clustering of vectors with high dimensionality, while their complexity grows exponentially with the acquisition size. We thus do not apply them to the datasets that have a large $\Dpool$. More details can be found in the Appendix~\ref{sec:efficiency}}.
We also evaluate a recently introduced cold-start acquisition function called \alps{}~\cite{yuan-etal-2020-cold} that uses the masked language model (MLM) loss of \bert{} 
as a proxy for model uncertainty in the downstream classification task. Specifically, aiming to leverage both uncertainty and diversity, \alps{}
forms a \textit{surprisal embedding} $s_x$ for each $x$, by passing the unmasked input $x$ through the \bert{} MLM head
to compute the cross-entropy loss for a random 15\% subsample of tokens against the target labels. 
\alps{} clusters these embeddings to sample $b$ sentences for each AL iteration. 
Lastly, we include \rand{}, that samples data from the pool from a uniform distribution.

\subsection{Implementation Details}
We use \textsc{BERT-base}~\cite{Devlin2019-ou} adding a task-specific classification layer using the implementation from the HuggingFace library~\cite{wolf-etal-2020-transformers}.
We evaluate the model $5$ times per epoch on 
the development set following \citet{Dodge2020FineTuningPL} and keep the one with the lowest validation loss.
We use the standard splits provided for all datasets, if available, otherwise we randomly sample a validation set from the training set. We test all models on a held-out test set.
We repeat all experiments with five different random seeds resulting into different initializations of the parameters of the model's extra task-specific output feedforward layer and the initial $\Dlab$.
For all datasets we use as budget the $15\%$ of $\Dpool$, initial training set $1\%$ and acquisition size $b=2\%$.
Each experiment is run on a single Nvidia Tesla V100 GPU.
More details are provided in the Appendix~\ref{sec:data_hyper}.

\begin{figure*}[!t]
    \begin{subfigure}{0.25\textwidth}
        \centering
        \includegraphics[width=\textwidth]{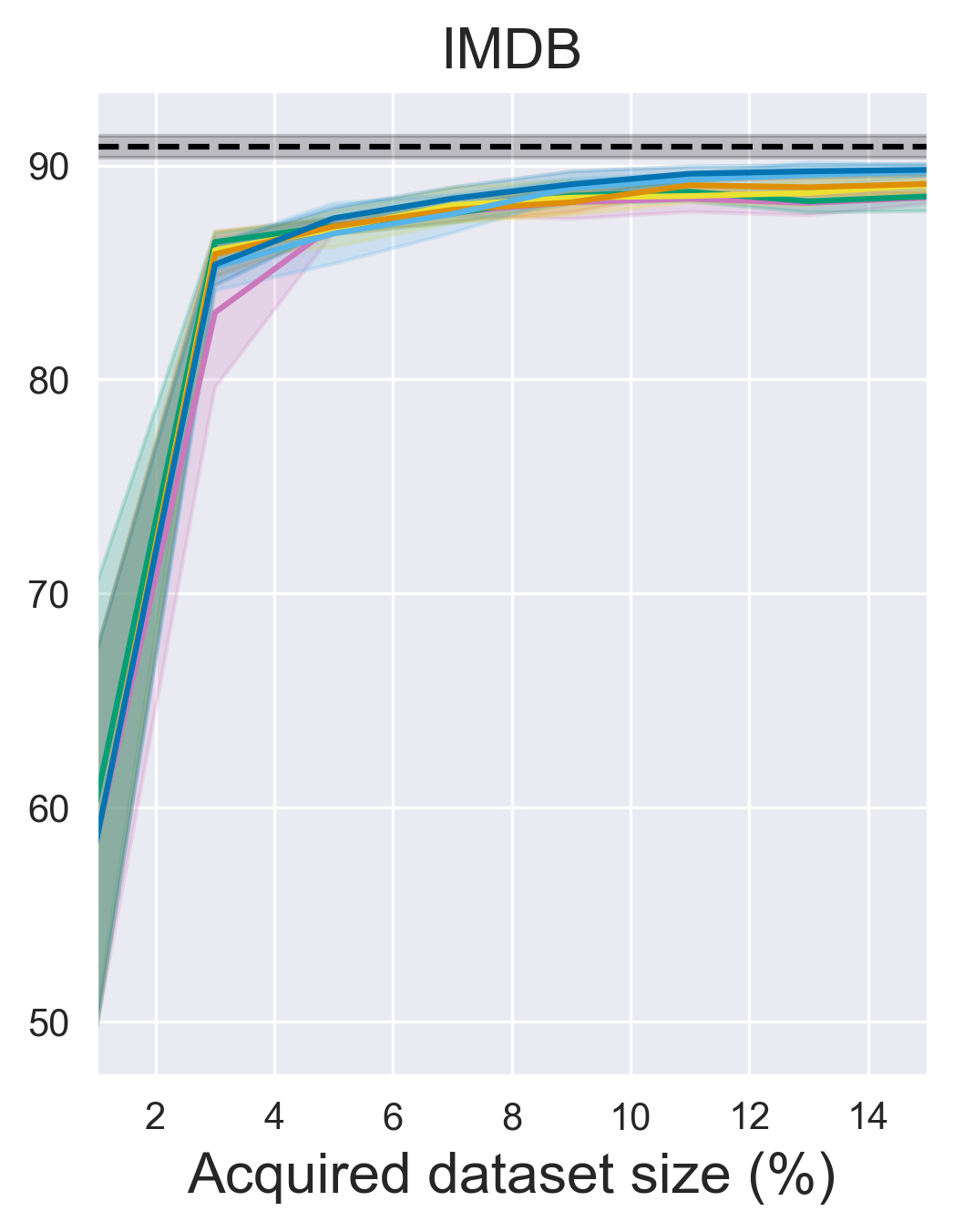}
    \end{subfigure}%
    \begin{subfigure}{0.25\textwidth}
        \centering
        \includegraphics[width=\textwidth]{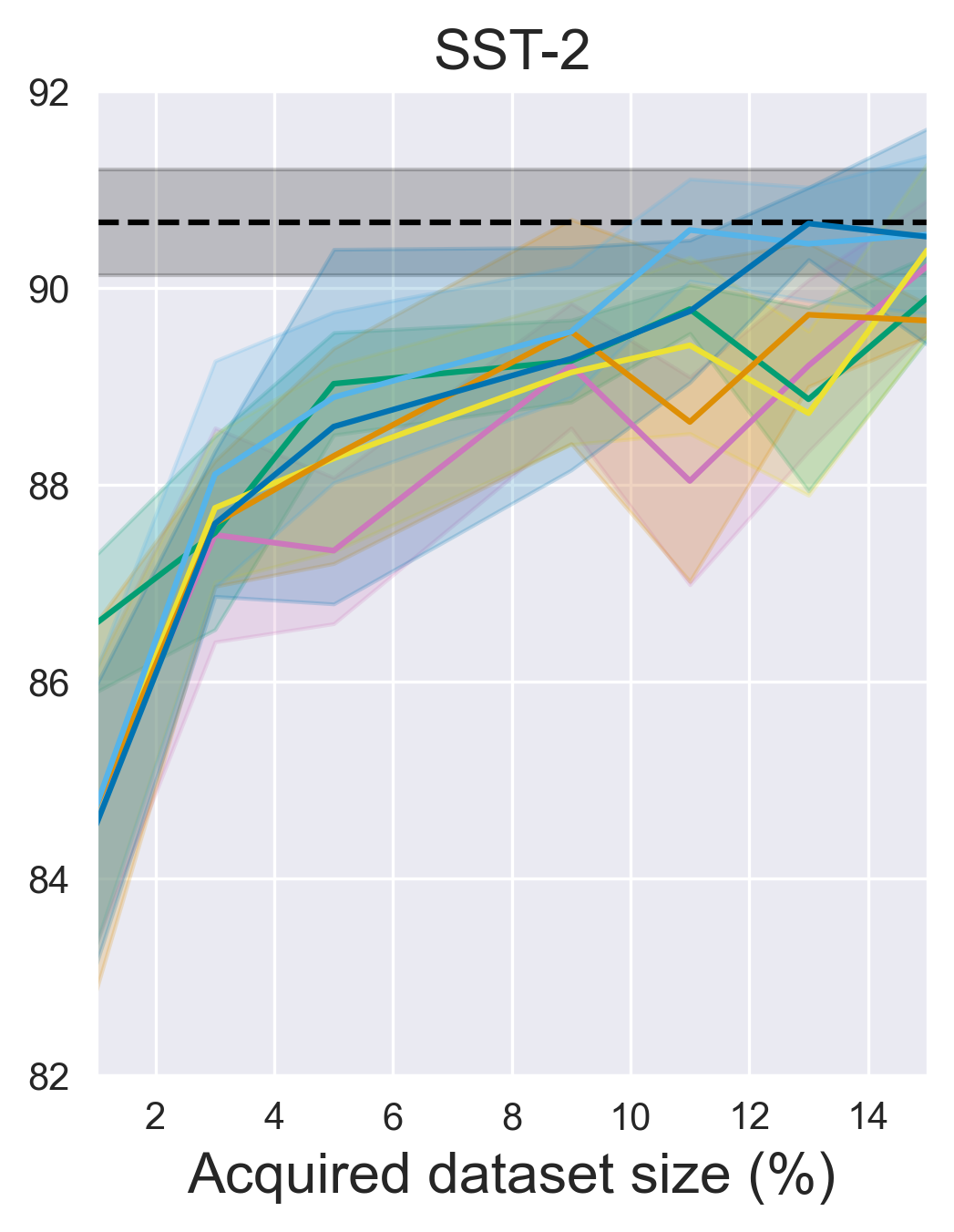}
    \end{subfigure}%
    \begin{subfigure}{0.25\textwidth}
        \centering
        \includegraphics[width=\textwidth]{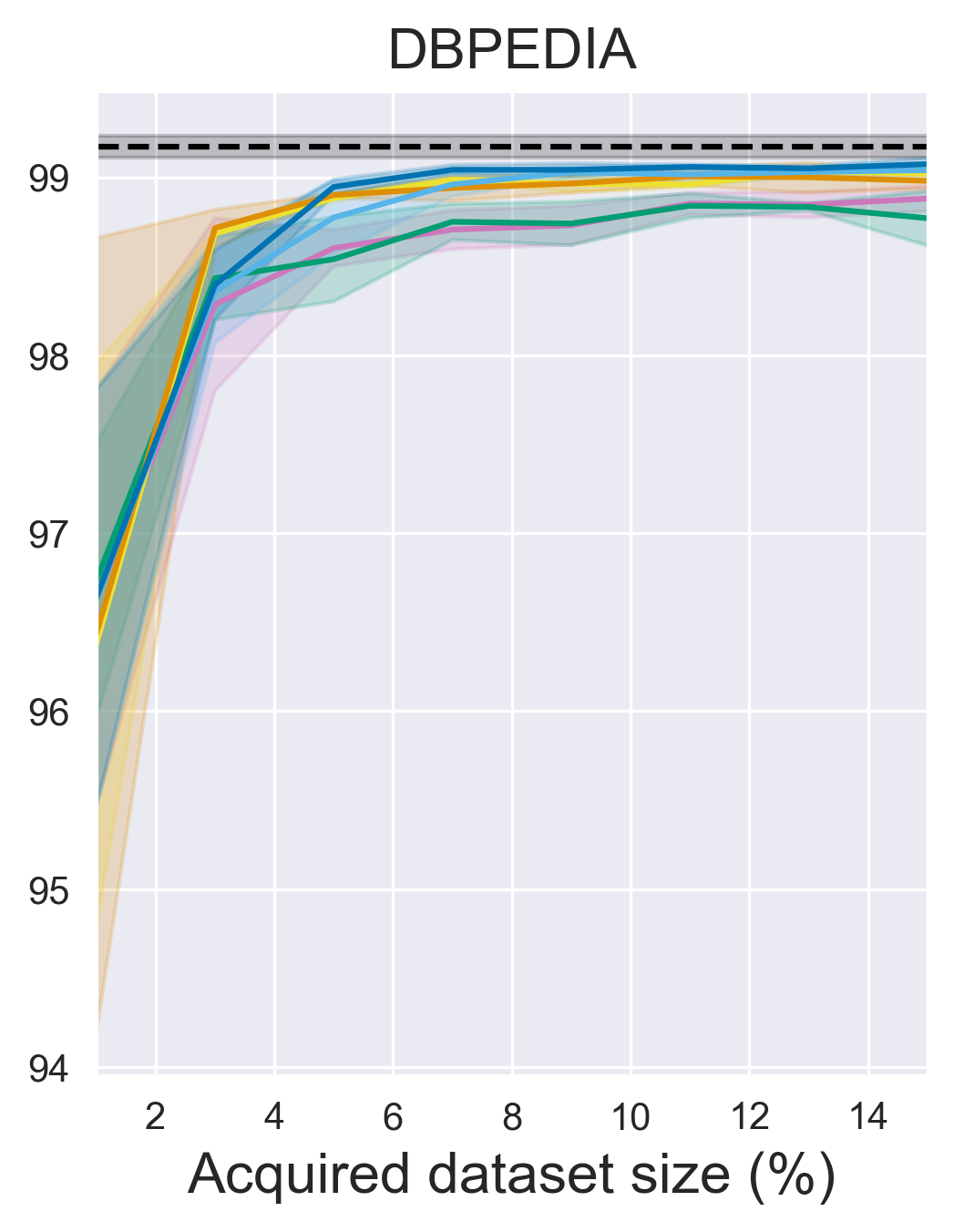}
    \end{subfigure}%
    \begin{subfigure}{0.25\textwidth}
        \centering
        \includegraphics[width=\textwidth]{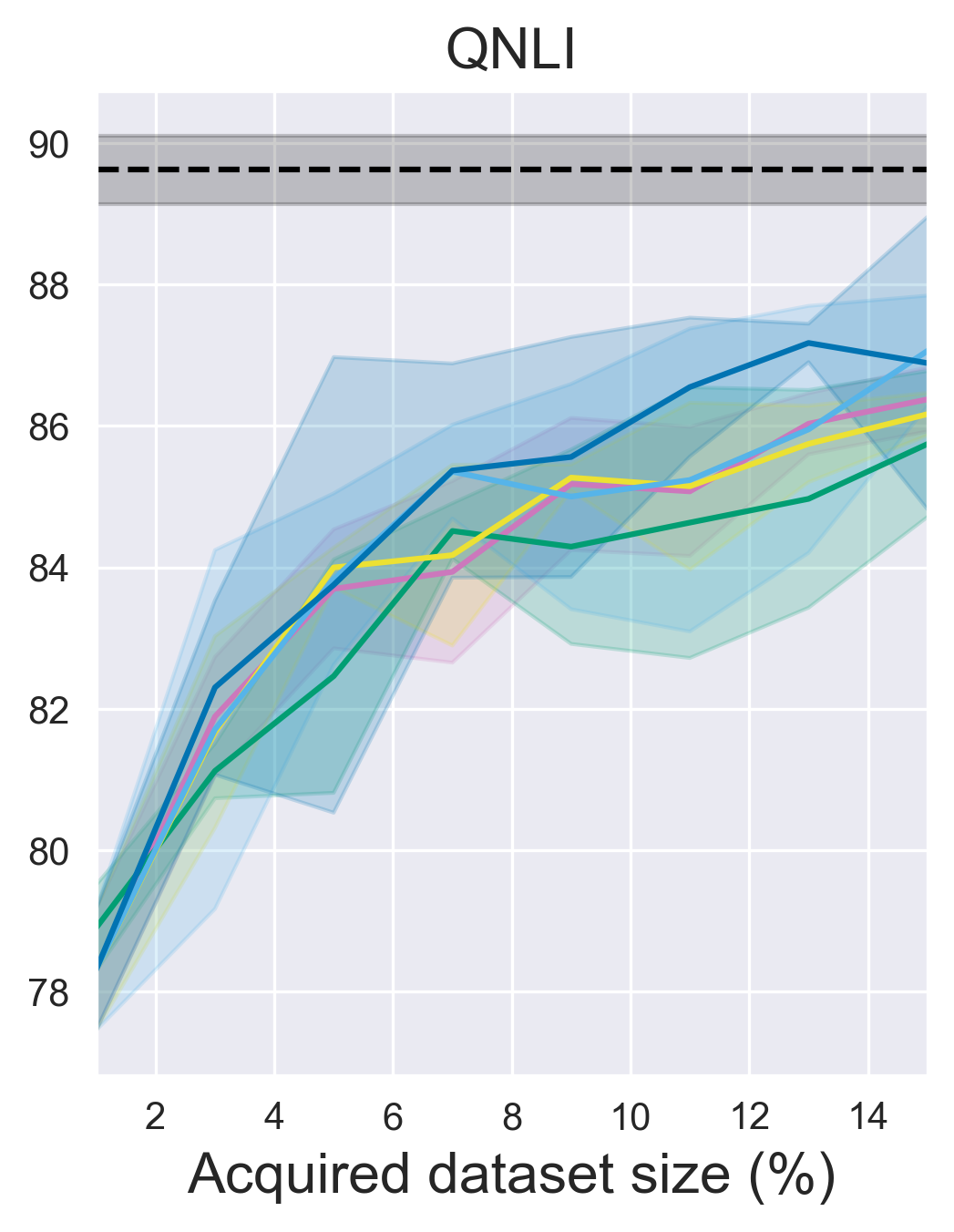}
    \end{subfigure}%
                \\
    \begin{subfigure}{0.25\textwidth}
        \centering
        \includegraphics[width=\textwidth]{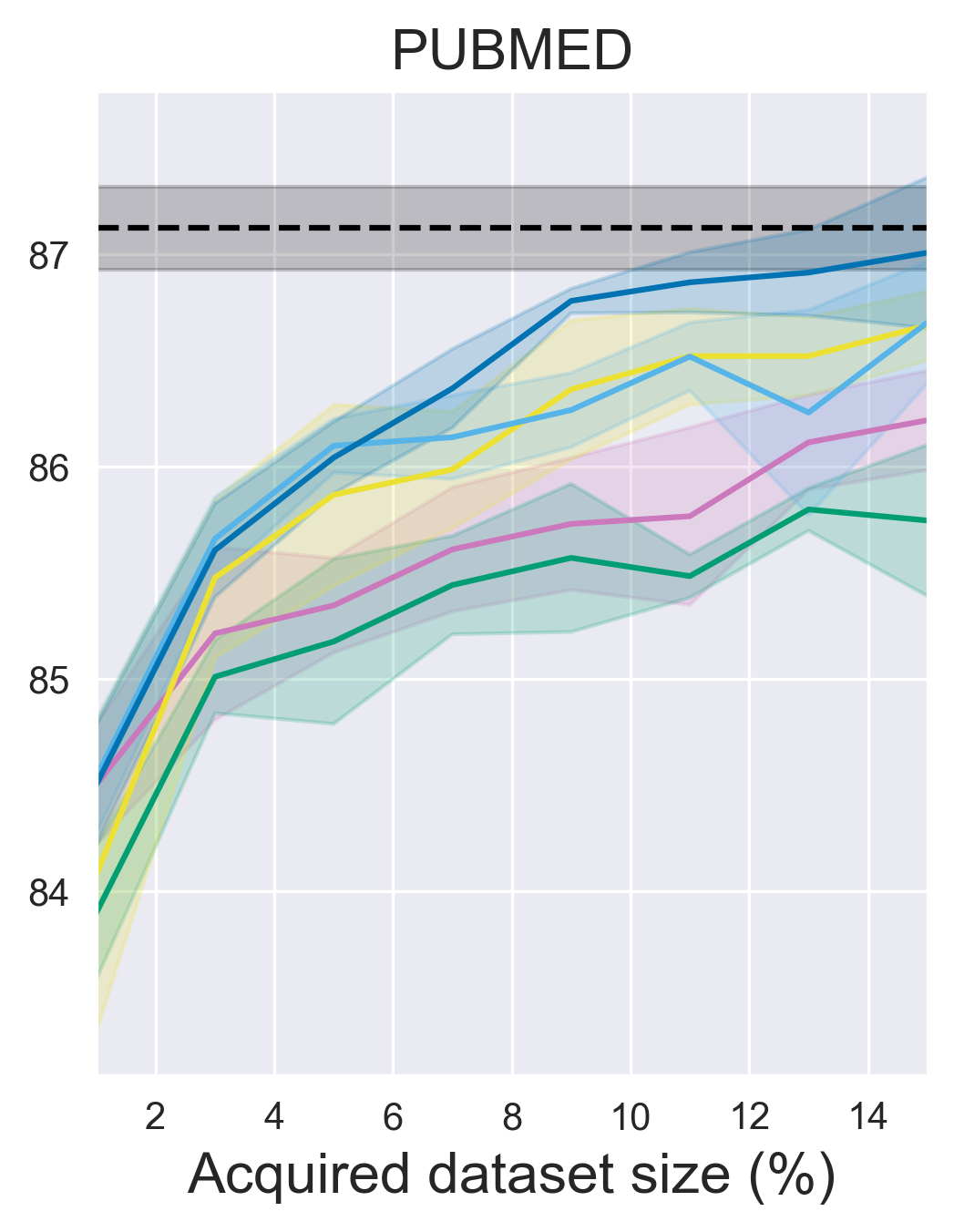}
    \end{subfigure}%
    \begin{subfigure}{0.25\textwidth}
        \includegraphics[width=\textwidth]{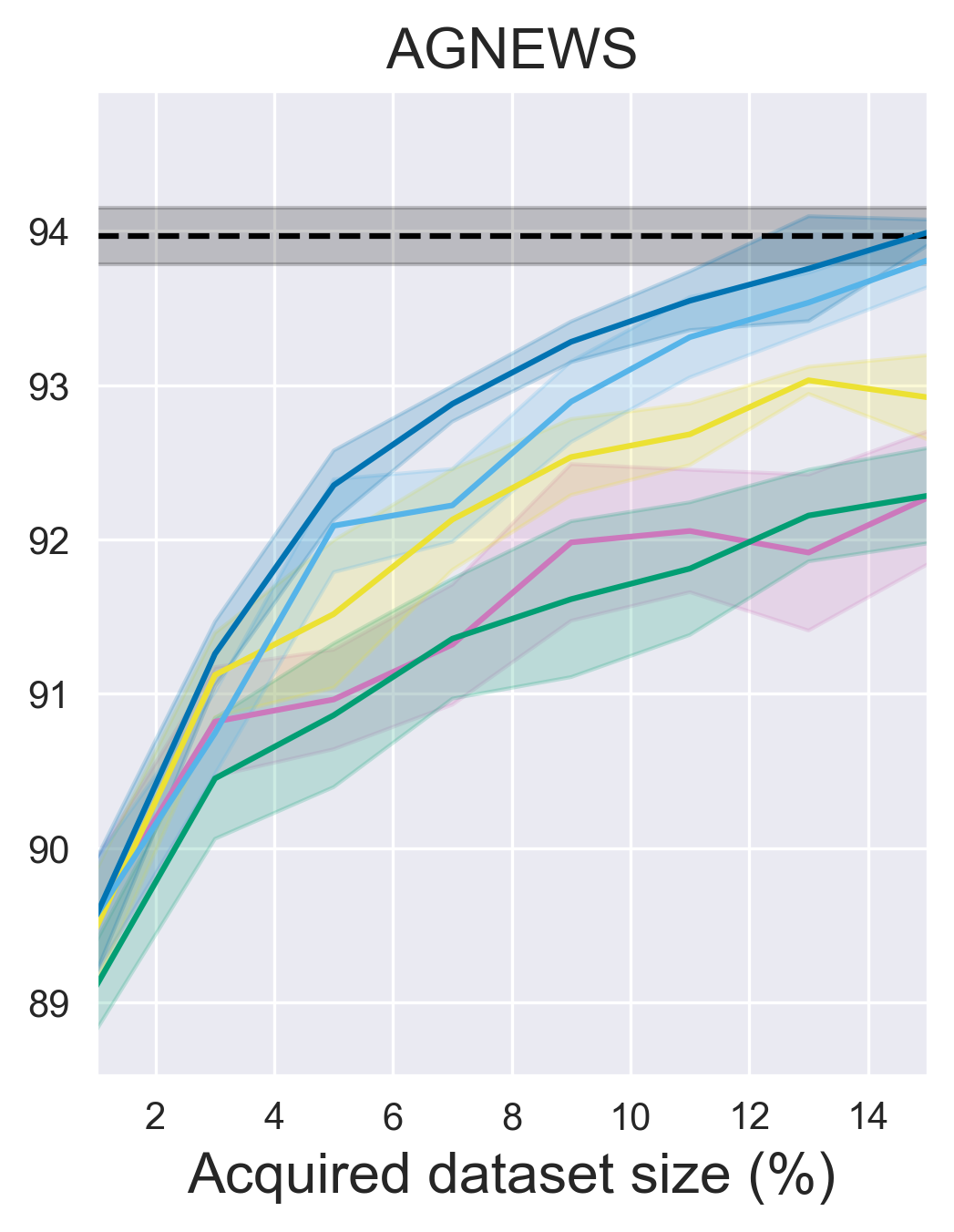}
    \end{subfigure}%
    \begin{subfigure}{0.25\textwidth}
        \centering
        \includegraphics[width=\textwidth]{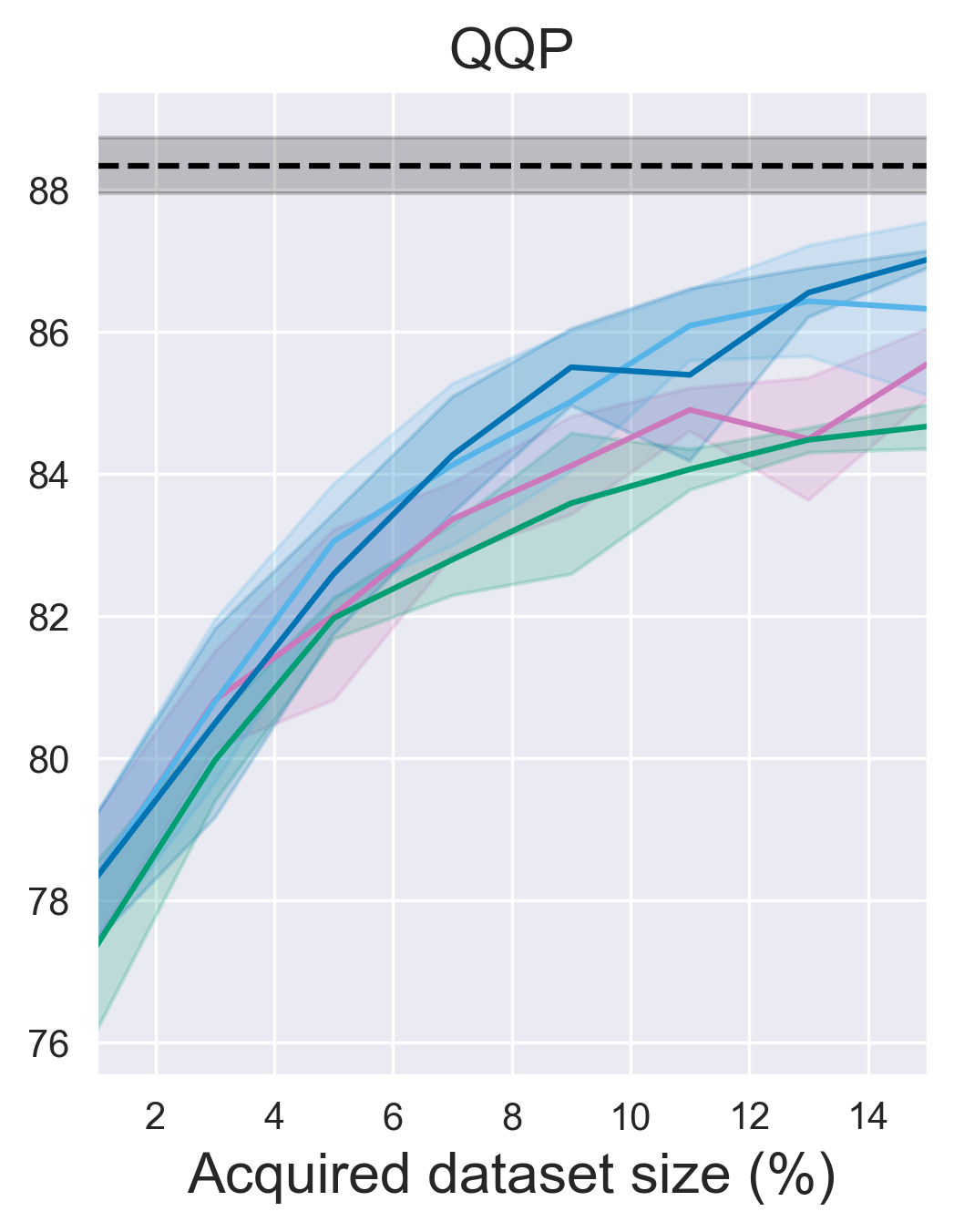}
    \end{subfigure}%
    \begin{subfigure}{0.25\textwidth}
        \centering
        \includegraphics[width=0.7\textwidth]{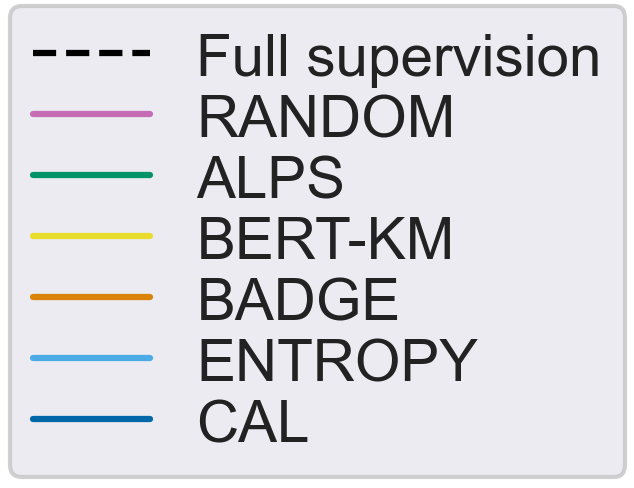}
    \end{subfigure}%
    \caption{In-domain (ID) test accuracy during AL iterations for different acquisition functions. 
    }
    \label{fig:al_id}
\end{figure*}

\section{Results}\label{sec:results}
\subsection{In-domain Performance}
We present results for in-domain test accuracy across all datasets and acquisition functions in Figure~\ref{fig:al_id}. 
We observe that \call{} is consistently the top performing method especially in \dbpedia{}, \pub{} and \ag{} datasets.

\call{} performs slightly better than \entropy{} in \imdb{}, \qnli{} and \qqp{}, while in \sst{} most methods yield similar results. 
\entropy{} is the second best acquisition function overall, consistently performing better than diversity-based or hybrid baselines. 
This corroborates recent findings from ~\citet{Desai2020-ys} that \bert{} is sufficiently calibrated (i.e. produces good uncertainty estimates), making it a tough baseline to beat in AL.

\bkm{} is a competitive baseline (e.g. \sst{}, \qnli{}) but always underperforms compared to \call{} and \entropy{}, suggesting that uncertainty is the most important signal in the data selection process. An interesting future direction would be to investigate in depth whether and which (i.e. which layer) representations of the current (pretrained language models) works best with similarity search algorithms and clustering.  

Similarly, we can see that \badge{}, despite using both uncertainty and diversity, also achieves low performance, indicating that clustering the constructed gradient embeddings does not benefit data acquisition.   
Finally, we observe that \alps{} generally underperforms and is close to \rand{}. We can conclude that this heterogeneous approach to uncertainty, i.e. using the pretrained language model as proxy for the downstream task, is beneficial only in the first few iterations, as shown in \citet{yuan-etal-2020-cold}. 

Surprisingly, we observe that for the \sst{} dataset \alps{} performs similarly with the highest performing acquisition functions, \call{} and \entropy{}. We hypothesize that due to the informal textual style of the reviews of \sst{} (noisy social media data), the pretrained \bert{} model can be used as a signal to query linguistically hard examples, that benefit the downstream sentiment analysis task. 
This is an interesting finding and a future research direction would be to investigate the correlation between the difficulty of an example in a downstream task with its perplexity (loss) of the pretrained language model.

\setlength{\tabcolsep}{6pt} 
\renewcommand{\arraystretch}{1.1} 

\begin{table}[!t]
\resizebox{\columnwidth}{!}{%
\centering
\small
\begin{tabular}{lccc}
\Xhline{2\arrayrulewidth}

       \textsc{train (id)}               & \sst{}   & \imdb & \textsc{qqp} \\
              \textsc{test (ood)}               & \imdb{}   & \sst{} & \twit{} \\\hline 
     \rand{} &  $76.28 \pm 0.72$ &  $82.50 \pm 3.61$ &  $85.86 \pm 0.48$ \\
   \bkm{} &  $75.99 \pm 1.01$ &  $84.98 \pm 1.22$ &     - \\
    \entropy{} &  $75.38 \pm 2.04$ &  $\textbf{85.54} \pm 2.52$ &  $85.06 \pm 1.96$ \\
       \alps{} &  $77.06 \pm 0.78$ &  $83.65 \pm 3.17$ &  $84.79 \pm 0.49$ \\
      \badge{} &     $76.41 \pm 0.92$ &  $85.19 \pm 3.01$ &     - \\
  \call{} &  $\textbf{79.00} \pm 1.39$ &  $84.96 \pm 2.36$ &  $\textbf{86.20} \pm 0.22$ \\
   \Xhline{2\arrayrulewidth}
\end{tabular}
}
\caption{Out-of-domain (OOD) accuracy of models trained with the actively acquired datasets created with different AL acquisition strategies.}

\label{table:ood}
\end{table}
\subsection{Out-of-domain Performance}
We also evaluate the out-of-domain (OOD) robustness of the models trained with the actively acquired datasets of the last iteration (i.e. $15\%$ of $\Dpool$ or 100\% of the AL budget) using different acquisition strategies.
We present the OOD results for \sst{}, \imdb{} and \qqp{} in 
Table~\ref{table:ood}.
When we test the models trained with \sst{} on \imdb{} (first column) we observe that \call{} achieves the highest performance compared to the other methods by a large margin, indicating that acquiring contrastive examples can improve OOD generalization. In the opposite scenario (second column), we find that the highest accuracy is obtained with \entropy{}. However, similarly to the ID results for \sst{} (Figure~\ref{fig:al_id}), all models trained on different subsets of the \imdb{} dataset result in comparable performance  when tested on the small \sst{} test set (the mean accuracies lie inside the standard deviations across models). 
We hypothesize that this is because \sst{} is not a challenging OOD dataset for the different \imdb{} models.
This is also evident by the high OOD accuracy, 85\% on average, which is close to the 91\% \sst{} ID accuracy of the full model (i.e. trained on 100\% of the ID data). 
Finally, we observe that \call{} obtains the highest OOD accuracy for \qqp{} compared to \rand{}, \entropy{} and \alps{}. 
Overall, our empirical results show that the models trained on the actively acquired dataset with \call{} obtain consistently similar or better performance than all other approaches when tested on OOD data.

\section{Ablation Study}\label{sec:ablation}
\begin{figure}[!t]
        \centering
        \includegraphics[width=\columnwidth]{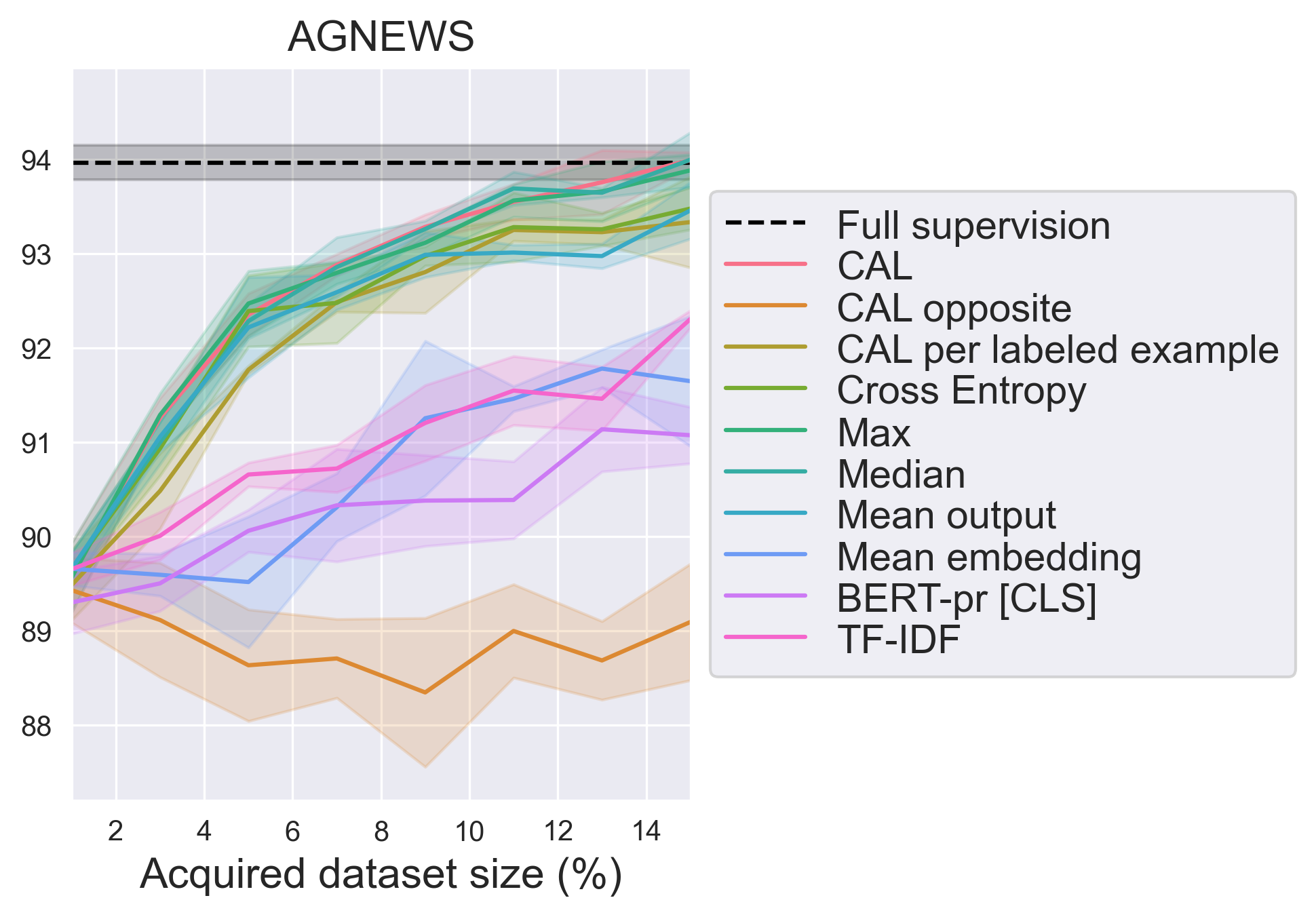}

    \caption{In-domain (ID) test accuracy with different variants of \call{} (ablation).}
    \label{fig:ablation}
\end{figure}

We conduct an extensive ablation study in order to provide insights for the behavior of every component of \call{}. 
We present all AL experiments on the \ag{} dataset in Figure~\ref{fig:ablation}.

\paragraph{Decision Boundary} We first aim to evaluate our hypothesis that \call{} acquires difficult examples that lie close to the model's decision boundary. Specifically, to validate that the ranking of the constructed neighborhoods is meaningful, we run an experiment where we acquire candidate examples that have the \textit{minimum} divergence from their neighbors opposite to \call{} (i.e. we replace $\mathrm{argmax}(.)$ with $\mathrm{argmin}(.)$ in line 8 of Algorithm \ref{algo:contrastive}). We observe
(Fig.~\ref{fig:ablation} - \texttt{CAL opposite}) that even after acquiring 15\% of unlabeled data, the performance remains unchanged compared to the initial model (of the first iteration), even degrades. In effect, this finding denotes that \call{} does select informative data points.

\paragraph{Neighborhood}
Next, we experiment with changing the way we construct the neighborhoods, aiming to improve computational efficiency. We thus modify our algorithm to create a neighborhood for each \textit{labeled} example (instead of unlabeled).\footnote{In this experiment, we essentially change the \textit{for-loop} of Algorithm~\ref{algo:contrastive} (cf. line 1-7) to iterate for each $x_l$ in $\Dlab{}$ (instead of each $x_p$ in $\Dpool{}$) and similarly find the $k$ nearest neighbors of each labeled example in the pool (KNN$(x_l, \Dpool{}, k)$) 
As for the scoring (cf. line 6), if an unlabeled example was not picked (i.e. was not a neighbor to a labeled example), its score is zero. If it was picked multiple times we average its scores. We finally acquire the top $b$ unlabeled data with the highest scores.
This formulation is more computationally efficient since usually $|\Dlab{}| << |\Dpool{}|$.}. This way we compute a divergence score only for the neighbors of the training data points.
However, we find this approach to slightly underperform (Fig.~\ref{fig:ablation} - \texttt{CAL per labeled example}), possibly because only a small fraction of the pool is considered and thus the uncertainty of all the unlabeled data points is not taken into account.

\paragraph{Scoring function}
We also experiment with several approaches for constructing our scoring function (cf. line 6 in Algorithm~\ref{algo:contrastive}).
Instead of computing the KL divergence between the predicted probabilities of each candidate example and its labeled neighbors (cf. line 5), we used cross entropy between the output probability distribution and the gold labels of the labeled data.
The intuition is to evaluate whether information of the actual label is more useful than the model's predictive probability distribution.
We observe this scoring function to result in a slight drop in performance (Fig.~\ref{fig:ablation} - \texttt{Cross Entropy}).
We also experimented with various pooling operations to aggregate the KL divergence scores for each candidate data point. We found maximum
and
median
(Fig.~\ref{fig:ablation} - \texttt{Max/Median})
to perform similarly with the average (Fig.~\ref{fig:ablation} - \texttt{\call{}}), which is the pooling operation we decided to keep in our proposed algorithm.

\paragraph{Feature Space}
Since our approach is related to to acquiring data near the model's decision boundary, this effectively translates into using the \texttt{[CLS]} output embedding of \bert{}. Still, we opted to cover several possible alternatives to the representations, i.e. feature space, that can be used to find the neighbors with KNN. We divide our exploration into two categories: intrinsic representations from the current fine-tuned model and extrinsic using different methods. For the first category, we examine representing each example with the mean embedding layer of \bert{} (Fig.~\ref{fig:ablation} -  \texttt{Mean embedding}) or the mean output embedding (Fig.~\ref{fig:ablation} - \texttt{Mean output}). We find both alternatives to perform worse than using the \texttt{[CLS]} token (Fig.~\ref{fig:ablation} - \texttt{\call{}}). The motivation for the second category is to evaluate whether acquiring contrastive examples in the \textit{input} feature space, i.e. representing the raw text, is meaningful \cite{gardner-etal-2020-evaluating}~\footnote{This can be interpreted as comparing the effectiveness of selecting data near the \textit{model} decision boundary vs. the \textit{task} decision boundary, i.e. data that are similar for the task itself or for the humans (in terms of having the same raw input/vocabulary), but are from different classes.}. We thus examine contextual representations from a pretrained \bert{} language model (Fig.~\ref{fig:ablation} - \texttt{BERT-pr [CLS]}) (not fine-tuned in the task or domain) and non-contextualized \textsc{tf-idf} vectors (Fig.~\ref{fig:ablation} - \texttt{TF-IDF}). We find both approaches, along with \texttt{Mean embedding}, to largely underperform compared to our approach that acquires ambiguous data near the model decision boundary. 



\section{Analysis}\label{sec:analysis}
Finally, we further investigate \call{} and all acquisition functions considered (baselines), in terms of diversity, representativeness and uncertainty.
Our aim is to provide insights on what data each method tends to select and what is the uncertainty-diversity trade-off of each approach. Table~\ref{table:unc_div} shows the results of our analysis averaged across datasets. We denote with $L$ the labeled set, $U$ the unlabeled pool and $Q$ an acquired batch of data points from $U$~\footnote{In the previous sections we used $\Dlab{}$ and $\Dpool{}$ to denote the labeled and unlabeled sets and we change the notation here to $L$ and $U$, respectively, for simplicity.}.

\subsection{Diversity \& Uncertainty Metrics}
\paragraph{Diversity in input space (\textsc{Div.-I})} We first evaluate the diversity of the actively acquired data in the input feature space, i.e. raw text, by measuring the overlap between tokens in the sampled sentences $Q$ and tokens from the rest of the data pool $U$. 
Following \citet{yuan-etal-2020-cold}, we compute \textsc{Div.-I} as the Jaccard similarity between the set of tokens from the sampled sentences $Q$, $\mathcal{V_Q}$, and the set of tokens from the
unsampled sentences $\mathcal{U \backslash Q}$, $\mathcal{V_{Q'}}$,
$\mathcal{J}(\mathcal{V}_\mathcal{Q}, \mathcal{V}_{\mathcal{Q}'}) = \frac{|\mathcal{V_Q} \cap \mathcal{V}_{\mathcal{Q}'}|}{|\mathcal{V_Q} \cup \mathcal{V}_{\mathcal{Q}'}|}$.
A high \textsc{Div.-I} value indicates high diversity because the sampled and unsampled sentences have many
tokens in common. 

\paragraph{Diversity in feature space (\textsc{Div.-F})}
We next evaluate diversity in the (model) feature space, using the \texttt{[CLS]} representations of a trained \bert{} 
model~\footnote{To enable an appropriate comparison, this analysis is performed after the initial \bert{} model is
trained with the initial training set and each AL strategy has selected examples equal to $2\%$ of the pool (first iteration). Correspondingly, all strategies
select examples from the same unlabeled set $U$
while using outputs from the same \bert{} model.}. 
Following \citet{Zhdanov2019-mg} and \citet{Ein-Dor2020-mm}, we compute \textsc{Div.-F} of a set $Q$ as
$\Big(\frac{1}{|U|} \sum\limits_{x_i \in U} \underset{x_j\in Q}{\mathrm{min }} d(\Phi(x_i), \Phi(x_j))
\Big)^{-1}$,
where $\Phi(x_i)$ denotes the  \cls{} output
token of example $x_i$ obtained by the model which was trained using $L$, and $d(\Phi(x_i), \Phi(x_j))$ denotes the Euclidean distance between $x_i$ and $x_j$ in the feature space.

\paragraph{Uncertainty (\textsc{Unc.})}
To measure uncertainty, we use the model $\mathcal{M}_{f}$ trained on the entire training dataset (Figure~\ref{fig:al_id} - \texttt{Full supervision}). 
As in \citet{yuan-etal-2020-cold}, we use the logits from
the fully trained model to estimate the uncertainty of an example, as it is a reliable estimate due to its high performance after training on
many examples,
while it
offers a fair comparison across all 
acquisition 
strategies.
First, we compute predictive
entropy of an input $x$ when evaluated by model $\mathcal{M}_{f}$ and then we take the average over all sentences in a sampled batch $Q$.
We use the average predictive entropy to estimate
uncertainty of the acquired batch $Q$ for each method
$-\frac{1}{|Q|} \sum\limits_{x \in Q} \sum\limits_{c=1}^C p(y=c|x)\mathrm{log}p(y=c|x)$. As a sampled batch $Q$ we use the full actively acquired dataset after completing our AL iterations (with $15\%$ of the data).

\paragraph{Representativeness (\textsc{Repr.})}
We finally analyze the representativeness of the acquired data as in~\citet{Ein-Dor2020-mm}.
We aim to study whether AL
strategies tend to select outlier examples that do not
properly represent the overall data distribution. We rely on the KNN-density measure proposed by \citet{zhu-etal-2008-active}, where the density of
an example is quantified by one over the average distance
between the example and its K most
similar examples (i.e., K nearest neighbors) within
$U$, based on the \cls{} representations as in \textsc{Div.-F}.
An example with high density degree is less likely to be an outlier. We define the representativeness
of a batch $Q$ as one over the average KNN-density
of its instances using the Euclidean distance with
K=$10$.

\setlength{\tabcolsep}{6pt} 
\renewcommand{\arraystretch}{1.1} 

\begin{table}[!t]
\resizebox{\columnwidth}{!}{%
\centering
\small
\begin{tabular}{lcccc}
\Xhline{2\arrayrulewidth}

                      & \textsc{Div.-I}  & \textsc{Div.-F} & \textsc{Unc.} & \textsc{Repr.}\\\hline 
     \rand{} &  0.766 &  0.356 &  0.132 &  1.848 \\
   \bkm{} &  0.717 &  \textbf{0.363} &  0.145 &  2.062 \\
    \entropy{} &  0.754 &  0.323 &  \textbf{0.240} &  2.442 \\
       \alps{} &  \textbf{0.771} &  0.360 &  0.126 &  2.038 \\
      \badge{} &  0.655 &  0.339 &  0.123 &  2.013 \\
  \call{} &  0.768 &  0.335 &  0.231 &  \textbf{2.693}\\ 
\Xhline{2\arrayrulewidth}
\end{tabular}
}
\caption{Uncertainty and diversity metrics across acquisition functions, averaged for all datasets.}

\label{table:unc_div}
\end{table}
\subsection{Discussion}
We first observe in Table~\ref{table:unc_div} that \alps{} acquires the most diverse data across all approaches. This is intuitive since \alps{} is the most linguistically-informed method as it essentially acquires data that are difficult for the language modeling task, thus favoring data with a more diverse vocabulary. All other methods acquire similarly diverse data, except \badge{} that has the lowest score. Interestingly, we observe a different pattern when evaluating diversity in the model feature space (using the \cls{} representations). \bkm{} has the highest \textsc{Div.-F} score, as expected, while \call{} and \entropy{} have the lowest. This supports our hypothesis that uncertainty sampling tends to acquire uncertain but similar examples, while \call{} by definition constrains its search in similar examples in the feature space that lie close to the decision boundary (contrastive examples). As for uncertainty, we observe that \entropy{} and \call{} acquire the most uncertain examples, with average entropy almost twice as high as all other methods. Finally, regarding representativeness of the acquired batches, we see that \call{} obtains the highest score, followed by \entropy{}, with the rest AL strategies to acquire less representative data. 

Overall, our analysis validates assumptions on the properties of data expected to be selected by the various acquisition functions. Our findings show that diversity in the raw text does not necessarily correlate with diversity in the feature space. In other words, low \textsc{Div.-F} does not translate to low diversity in the distribution of acquired tokens (\textsc{Div.-I}), suggesting that \call{} can acquire similar examples in the feature space that have sufficiently diverse inputs. Furthermore, combining the results of our AL experiments (Figure~\ref{fig:al_id}) and our analysis (Table~\ref{table:unc_div}) we conclude that the best performance of \call{}, followed by \entropy{}, is due to acquiring uncertain data. We observe that the most notable difference, in terms of selected data, between the two approaches and the rest is uncertainty (\textsc{Unc.}), suggesting perhaps the superiority of uncertainty over diversity sampling. We show that \call{} improves over \entropy{} because our algorithm ``guides'' the focus of uncertainty sampling by not considering redundant uncertain data that lie away from the decision boundary and thus improving representativeness.     
We finally find that \rand{} is evidently the worst approach, as it selects the least diverse and uncertain data on average compared to all methods.

\section{Related Work}\label{sec:background}

\paragraph{Uncertainty Sampling} Uncertainty-based acquisition for AL focuses on selecting data points that the model predicts with low confidence. A simple uncertainty-based acquisition function is \textit{least confidence} \cite{Lewis:1994:SAT:188490.188495} that sorts data in descending order from the pool by the probability of not predicting the most confident class.
Another approach is to select samples that maximize the predictive entropy. \citet{Houlsby2011-qz} propose Bayesian Active Learning by Disagreement (BALD), a method that chooses data points that maximize the mutual information between predictions and model's posterior probabilities. 
\citet{Gal2017-gh} applied BALD for deep neural models using Monte Carlo dropout~\cite{Gal2016-lf} to acquire multiple uncertainty estimates for each candidate example.
Least confidence, entropy and BALD acquisition functions have been applied in a variety of text classification and sequence labeling tasks, showing to substantially improve data efficiency
\cite{Shen2017-km,Siddhant2018-lg,
Lowell2019-mf,Kirsch2019-lk,
shelmanov-etal-2021-active, DBLP:journals/corr/abs-2104-08320}.

\paragraph{Diversity Sampling} On the other hand, diversity or representative sampling is based on selecting batches of unlabeled examples that are representative of the unlabeled pool, based on the intuition that a representative set of examples once labeled, can act as a surrogate for the full data available.  
In the context of deep learning,
 \citet{DBLP:journals/corr/abs-1711-00941} and \citet{conf/iclr/SenerS18} select representative examples based on core-set construction, a fundamental problem in computational geometry. Inspired by generative adversarial learning,
\citet{DBLP:journals/corr/abs-1907-06347} define AL as a binary classification task with
an adversarial classifier trained to not be able to discriminate data from the training set and the pool.
Other approaches based on adversarial active learning, 
use out-of-the-box models to perform adversarial attacks on the training data, in order to approximate the distance from the decision boundary of the model~\cite{Ducoffe2018-sq,ru-etal-2020-active}.

\paragraph{Hybrid} There are several existing approaches that combine representative and uncertainty sampling. 
Such approaches include active learning algorithms that use meta-learning ~\cite{10.5555/1005332.1005342,conf/aaai/HsuL15}
and reinforcement learning~\cite{fang-etal-2017-learning,liu-etal-2018-learning-actively}, aiming to learn a policy for switching between a diversity-based or an uncertainty-based criterion at each iteration.
Recently, \citet{Ash2020Deep} propose Batch Active learning by Diverse Gradient Embeddings (\badge{}) and \citet{yuan-etal-2020-cold} propose Active Learning by Processing Surprisal (\alps{}), a cold-start acquisition function specific for pretrained language models.
Both methods construct representations for the unlabeled data based on uncertainty, and then use them for clustering; hence combining both uncertainty and diversity sampling.
The effectiveness of AL in a variety of NLP tasks with pretrained language models, e.g. \bert{}~\cite{Devlin2019-ou}, has empirically been recently evaluated by \citet{Ein-Dor2020-mm}, showing substantial improvements over random sampling.

\section{Conclusion \& Future Work}
We present \call{}, a novel acquisition function for AL that acquires \textit{contrastive examples}; data points which are similar in the model feature space and yet the model outputs maximally different class probabilities. Our approach uses information from the feature space to create neighborhoods for each unlabeled example, and predictive likelihood for ranking the candidate examples. Empirical experiments on various in-domain and out-of-domain scenarios demonstrate that  \call{} performs  better than other acquisition functions in the majority of cases. After analyzing the actively acquired datasets obtained with all methods considered, we conclude that entropy is the hardest baseline to beat, but our approach improves it by guiding uncertainty sampling in regions near the decision boundary with more informative data. 

Still, our empirical results and analysis show that there is no single acquisition function to outperform all others consistently \textit{by a large margin}. This demonstrates that there is still room for improvement in the AL field.

Furthermore, recent findings show that in specific tasks, as in Visual Question Answering (VQA), complex acquisition functions might not outperform random sampling because they tend to select \textit{collective outliers} that hurt model performance~\cite{karamcheti-etal-2021-mind}. 
We believe that taking a step back and analyzing the behavior of standard acquisition functions, e.g. with Dataset Maps~\cite{swayamdipta-etal-2020-dataset}, might be beneficial. Especially, if similar behavior appears in other NLP tasks too.

Another interesting future direction for \call{}, related to interpretability, would be to evaluate whether acquiring contrastive examples for the \textit{task}~\cite{Kaushik2020Learning,gardner-etal-2020-evaluating} is more beneficial than contrastive examples for the \textit{model}, as we do in \call{}.

\section*{Acknowledgments}
KM and NA are supported by Amazon through the Alexa Fellowship scheme.


\bibliography{anthology,custom}
\bibliographystyle{acl_natbib}

\clearpage 
\appendix

\section{Appendix}
\label{sec:appendix}

\subsection{Data \& Hyperparameters}\label{sec:data_hyper}
In this section we provide details of all the datasets we used in this work and the hyperparparameters used for training the model. For \qnli{}, \imdb{} and \sst{} we randomly sample 10\% from the training set to serve as the validation set, while for \ag{} and \qqp{} we sample 5\%. For the \dbpedia{} dataset we undersample both training and validation datasets (from the standard splits) to facilitate our AL simulation (i.e. the original dataset consists of 560K training and 28K validation data examples). For all datasets we use the standard test set, apart from \sst{}, \qnli{} and \qqp{} datasets that are taken from the \textsc{glue} benchmark~\cite{wang2018glue} we use the development set as the held-out test set and subsample a development set from the training set.

For all datasets we train \textsc{BERT-base}~\cite{Devlin2019-ou} from the HuggingFace library~\cite{wolf-etal-2020-transformers} in Pytorch~\cite{NEURIPS2019_9015}. We train all models with batch size $16$, learning rate $2e-5$, no weight decay, AdamW optimizer with epsilon $1e-8$. For all datasets we use maximum sequence length of $128$, except for \imdb{} that contain longer input texts, where we use $256$. To ensure reproducibility and fair comparison between the various methods under evaluation, we run all experiments with the same five seeds that we randomly selected from the range $[1,9999]$.
We evaluate the model 5 times per epoch on 
the development set
 following \citet{Dodge2020FineTuningPL} and keep the one with the lowest validation loss.
We use the code provided by \citet{yuan-etal-2020-cold} for \alps{}, \badge{} and \bkm{}.

\subsection{Efficiency}\label{sec:efficiency}
In this section we compare the efficiency of the acquisition functions considered in our experiments.
We denote $m$ the number of labeled data in $\Dlab$, $n$ the number of unlabeled data in $\Dpool$, $C$ the number of classes in the downstream classification task, $d$ the dimension of embeddings,
$t$ is fixed number of iterations for k-MEANS, $l$ the maximum sequence length and $k$ the acquisition size.
In our experiments, following~\cite{yuan-etal-2020-cold}, $k=100$, $d=768$, $t=10$, and $l=128$\footnote{Except for \imdb{} where $l=256$.}.
\alps{} requires $\mathcal{O}(tknl)$ considering that the surprisal embeddings are computed.
\bkm{} and \badge{}, the most computationally heavy approaches, require $\mathcal{O}(knd)$ and $\mathcal{O}(Cknd)$ respectively, given that gradient embeddings are computed for \badge{}~\footnote{This information is taken from Section 6 of \citet{yuan-etal-2020-cold}.}.
On the other hand, \entropy{} only requires $n$ forward passes though the model, in order to obtain the logits for all the data in $\Dpool{}$.
Instead, our approach, \call{}, first requires $m+n$ forward passes, in order to acquire the logits and the \texttt{CLS} representations of the the data (in $\Dpool{}$ and $\Dlab{}$) and then 
one iteration for all data in $\Dpool$ to obtain the scores. 

We present the runtimes in detail for all datasets and acquisition functions in Tables~\ref{table:runtimes} and ~\ref{table:runtimes_avg}.
First, we define the total \textit{acquisition time} as a sum of two types of times; \textit{inference} and \textit{selection} time.
Inference time is the time that is required in order to pass all data from the model to acquire predictions or probability distributions or model encodings (representations). This is explicitly required for the uncertainty-based methods, like \entropy{}, and our method \call{}.
The remaining time is considered \textit{selection} and essentially is the time for all necessary computations in order to rank and select the $b$ most important examples from $\Dpool{}$.

We observe in Table~\ref{table:runtimes} that the diversity-based functions do not require this explicit inference time, while for \entropy{} it is the only computation that is needed (taking the argmax of a list of uncertainty scores is negligible). \call{} requires both inference and selection time. We can see that inference time of \call{} is a bit higher than \entropy{} because we do $m+n$ forward passes instead of $n$, that is equivalent to both $\Dpool{}$ and $\Dlab{}$ instead of only $\Dpool{}$. The selection time for \call{} is the \textit{for-loop} as presented in our Algorithm~\ref{algo:contrastive}. We observe that it is often less computationally expensive than the inference step (which is a simple forward pass through the model). Still, there is room for improvement in order to reduce the time complexity of this step. 

In Table~\ref{table:runtimes_avg} we present the total time for all datasets (ordered with increasing $\Dpool{}$ size) and the average time for each acquisition function, as a means to rank their efficiency. Because we do not apply all acquisition functions to all datasets we compute three different average scores in order to ensure fair comparison. \textsc{avg.-all} is the average time across all $7$ datasets and is used to compare \rand{}, \alps{}, \entropy{} and \call{}.
\textsc{avg.-3} is the average time across the first $3$ datasets (\imdb{}, \sst{} and \dbpedia{}) and is used to compare all acquisition functions.
Finally, \textsc{avg.-6} is the average time across all datasets apart from \qqp{} and is used to compare \rand{}, \alps{}, \bkm{}, \entropy{} and \call{}.

We first observe that \entropy{} is overall the most efficient acquisition function.
According to the \textsc{avg.-all} column, we observe that \call{} is the second most efficient function, followed by \alps{}. According to the \textsc{avg.-6} we observe the same pattern, with \bkm{} to be the slowest method. Finally, we compare all acquisition functions in the $3$ smallest (in terms of size of $\Dpool{}$) datasets and find that \entropy{} is the fastest method followed by \alps{} and \call{} that require almost 3 times more computation time. The other clustering methods, \bkm{} and \badge{}, are significantly more computationally expensive, requiring respectively 13 and 100(!) times more time than \entropy{}. 

\setlength{\tabcolsep}{6pt} 
\renewcommand{\arraystretch}{1.1} 

\begin{table*}[!t]
\resizebox{\textwidth}{!}{%
\centering
\begin{tabular}{lccccccc}
\Xhline{2\arrayrulewidth}

{} &      \dbpedia &        \imdb &        \sst{} &         \qnli &      \ag &         \pub &            \qqp \\
\midrule
\rand{}    &      $(0,0)$ &     $(0,0)$ &      $(0,0)$ &      $(0,0)$ &      $(0,0)$ &        $(0,0)$ &        $(0,0)$ \\
\alps{}     &    $(0,181)$ &   $(0,222)$ &    $(0,733)$ &   $(0,1607)$ &   $(0,2309)$ &     $(0,5878)$ &    $(0,14722)$ \\
\bkm        &    $(0,467)$ &   $(0,431)$ &   $(0,4265)$ &   $(0,8138)$ &   $(0,9344)$ &    $(0,25965)$ &   $(-,-)$ \\
\badge{}    &  $(0,12871)$ &  $(0,3816)$ &  $(0,25640)$ &      $(-,-)$ &      $(-,-)$ &        $(-,-)$ &        $(-,-)$ \\
\entropy{} &    $(103,1)$ &   $(107,0)$ &    $(173,0)$ &    $(331,0)$ &    $(402,0)$ &      $(596,0)$ &     $(1070,0)$ \\
\call{}    &   $(133,49)$ &  $(212,61)$ &  $(464,244)$ &  $(528,376)$ &  $(656,628)$ &  $(1184,1445)$ &  $(1541,2857)$ \\
\hline

\Xhline{2\arrayrulewidth}
\end{tabular}
}

\caption{Runtimes (in seconds) for all datasets and acquisition functions. In each cell of the table we present a tuple $(i,s)$ where $i$ is the \textit{inference time} and $s$ the \textit{selection time}. \textit{Inference time} is the time for the model to perform a forward pass for all the unlabeled data in $\Dpool$ and \textit{selection time} is the time that each acquisition function requires to rank all candidate data points and select $b$ for annotation (for a single iteration). Since we cannot report the runtimes for \textit{every} model in the AL pipeline (at each iteration the size of $\Dpool$ changes), we provide the median.}

\label{table:runtimes}
\end{table*}

\setlength{\tabcolsep}{6pt} 
\renewcommand{\arraystretch}{1.1} 

\begin{table*}[!t]
\resizebox{\textwidth}{!}{%
\centering
\begin{tabular}{lccccccc|ccc}
\Xhline{2\arrayrulewidth}

{} &      \dbpedia &        \imdb &        \sst{} &         \qnli &      \ag &         \pub &            \qqp  & \textsc{avg.-all}& \textsc{avg.-3} & \textsc{avg.-6}\\
\midrule
\rand{}    &      $0$ &     $0$ &      $0$ &     $0$ &     $0$ &      $0$ &       $0$ &        $0$ &                   $0$ &                   $0$ \\
\alps{}     &    $181$ &   $222$ &    $733$ &  $1607$ &  $2309$ &   $5878$ &   $14722$ &     $3664$ &                 $378$ &                $1821$ \\
\bkm        &    $467$ &   $431$ &   $4265$ &  $8138$ &  $9344$ &  $25965$ &  $-$ &    $-$ &                $1721$ &                $8101$ \\
\badge{}    &  $12871$ &  $3816$ &  $25640$ &     $-$ &     $-$ &      $-$ &       $-$ &     $-$ &               $14109$ &                $-$ \\
\entropy{} &    $104$ &   $107$ &    $173$ &   $331$ &   $402$ &    $596$ &    $1070$ &      $397$ &                 $128$ &                 $285$ \\
\call{}    &    $182$ &   $273$ &    $708$ &   $904$ &  $1284$ &   $2629$ &    $4398$ &     $1482$ &                 $387$ &                 $996$ \\
\hline

\Xhline{2\arrayrulewidth}
\end{tabular}
}

\caption{Runtimes (in seconds) for all datasets and acquisition functions. In each cell of the table we present the total acquisition time (inference and selection). \textsc{avg.-all} shows the average acquisition time for each acquisition function for all datasets, \textsc{avg.-6}. for all datasets except \qqp{} and \textsc{avg.-3} for the $3$ first datasets only (\dbpedia{}, \imdb{}, \sst{}).}

\label{table:runtimes_avg}
\end{table*}

Interestingly, we observe the effect of the acquisition size ($2\%$ of $\Dpool{}$ in our case) and the size of $\Dpool{}$ in the clustering methods. As these parameters increase, the computation of the corresponding acquisition function increases dramatically. For example, we observe that in the $3$ smallest datasets that \alps{} requires similar time to \call{}. However, when we increase $b$ and $m$ (i.e. as we move from \dbpedia{} with $20K$ examples in $\Dpool{}$ to \qnli{} with $100K$ etc - see Table~\ref{table:datasets}) we observe that the acquisition time of \alps{} becomes twice as much as that of \call{}. For instance, in \qqp{} with acquisition size $3270$ we see that \alps{} requires $14722$ seconds on average, while \call{} $4398$. This shows that even though our approach is more computationally expensive as the size of $\Dpool{}$ increases, the complexity is linear, while for the other hybrid methods that use clustering, the complexity grows exponentially.

\subsection{Reproducibility}
All code for data preprocessing, model implementations, and active learning algorithms is made available at {\small \url{https://github.com/mourga/contrastive-active-learning}}. 
For questions regarding the implementation, please contact the first author.

\end{document}